\documentclass[11pt]{article}
\usepackage[margin=1in]{geometry}
\usepackage[T1]{fontenc}
\usepackage[utf8]{inputenc}
\usepackage{microtype}
\usepackage{amsmath,amssymb,amsthm}
\usepackage{graphicx,booktabs,multirow,makecell,tabularx,array}
\usepackage{siunitx}
\usepackage{xurl}
\usepackage{placeins}
\usepackage{enumitem}
\usepackage[round,authoryear]{natbib}
\usepackage[hidelinks]{hyperref}
\newcommand{\MARC}{\textsc{MARC}}
\newcommand{\MARCP}{\textsc{MARC-P}}
\newcommand{\MARCS}{\textsc{MARC-S}}

\newtheorem{theorem}{Theorem}
\newtheorem{corollary}[theorem]{Corollary}
\theoremstyle{remark}\newtheorem{remark}[theorem]{Remark}
\title{When the Cheapest Feasible Router Fails:\\Auditing Deployment Commitments in Cost-Aware LLM Routing}
\author{Minxuan Hu$^{1,*}$, Qishi Zhan$^{2,*,\dagger}$, Ziheng Chen$^{3}$,\\
Zhengtao Yao$^{4}$, Zihan Dong$^{5}$,\\
Jiayu Yi$^{6}$, Wenxi Sun$^{7}$\\[0.6em]
\small $^{1}$Cornell University \quad $^{2}$Marquette University\\
\small $^{3}$University of Texas at Austin \quad $^{4}$University of Southern California\\
\small $^{5}$Georgia Institute of Technology \quad $^{6}$Nanyang Technological University\\
\small $^{7}$Johns Hopkins University\\[0.4em]
\small $^*$Equal contribution. $^\dagger$Corresponding author: \texttt{qishi.zhan@marquette.edu}}
\date{}
\begin{document}
\maketitle
\begin{abstract}
Cost-aware LLM routing often selects the cheapest policy meeting empirical risk and coverage constraints. Yet a router that passes an audit can violate its fixed service contract on new requests, even when candidate policies are constructed independently. We study this deployment-commitment decision with disjoint construction, audit, and test samples. Across 19 RouterBench task families, the cheapest audit-feasible router deploys in 91.82\% of opportunities but violates the held-out contract in 32.22\% of deployments. We introduce Menu-Aware Routing Commitment (\MARC), which qualifies policies using audit evidence and a serving-cost tolerance to retain the apparent winner, deploy a supported alternative, or defer. Its practical policywise variant (\MARCP) reduces observed conditional failure to 0.73\% at 85.79\% deployment, with higher serving cost; fixed-margin and formal risk-control procedures provide competitive alternatives. Nested audits show how added evidence changes deployable policy sets, and learned semantic routing exhibits the same commitment failure. These results establish deployment commitment as a distinct operational decision connecting router evaluation, service reliability, and inference cost.
\end{abstract}

\section{Introduction}\label{sec:introduction}
Cost-aware language-model routing aims to direct each request to an economical model while preserving useful response quality \citep{chen2023frugalgpt,ding2024hybrid,ong2024routellm,somerstep2025carrot}. Selective routers may also decline to answer, trading model-call cost against the fraction of requests served. Service-aware routing and calibrated selection methods increasingly express such decisions through reliability requirements \citep{woisetschlager2025mess,hao2026racer,uddin2026conformal}. Consider an operator requiring at most $r^\star$ accepted-query loss and at least $c^\star$ coverage. Among evaluated routing configurations, choosing the least expensive policy that meets both measured thresholds is a natural recommendation. Yet the router that looks cheapest and feasible on an audit sample may not satisfy the promised contract on new traffic.

The deployment problem does not end when a router has been constructed. Given a menu of candidate policies specifying model assignments or abstention, an operator must decide which policy has enough evidence to serve traffic under a fixed contract. Cost optimization can favor weaker model choices or lower acceptance, leaving little headroom against the risk ceiling or coverage floor. Sampling fluctuations can then make boundary-adjacent policies appear feasible, while selecting the cheapest among them favors optimistic estimates. These effects connect classical model-selection bias and high-confidence policy evaluation \citep{cawley2010overfitting,thomas2015highconfidence,angelopoulos2022learn,laufer2023pareto} to a concrete routing decision: which policy serves requests, how many requests it answers, and at what inference cost. Even with a policy menu frozen independently of the audit, the final point-estimate feasibility check remains a consequential source of deployment error.

We study this decision under prespecified risk--coverage contracts and disjoint construction, audit, and test partitions. On RouterBench, the cheapest independently audited feasible router deploys in 91.82\% of opportunities yet violates its held-out contract in 32.22\% of deployments. Even selecting a router before auditing and applying a separate point-estimate gate yields 21.08\% conditional failure at 63.00\% deployment. The violations have meaningful magnitude: among failed winner deployments, mean severity is 3.15 percentage points. A separately fitted CARROT-style semantic-routing family shows the same deployment vulnerability. Thus the problem persists beyond reused construction data and beyond hand-designed routing policies.

We introduce \emph{Menu-Aware Routing Commitment} (\MARC) to turn independent audit evidence into an operational decision over a frozen policy menu. MARC first identifies policies supported under the fixed contract, then selects a cost-tolerant alternative with stronger service-support margins. It can \emph{retain} the cheapest apparent winner, \emph{replace} that winner with a supported policy, or \emph{defer} deployment. Two evidence constructions instantiate this interface: practical policywise bootstrap adjustments with explicit margins (\MARCP) and a simultaneous-menu correction based on shared resampling (\MARCS). We evaluate their deployment consequences alongside fixed buffers, Learn-then-Test, Pareto Testing, and exact controls.

Our contributions are threefold. \emph{First}, we characterize how independently audited, cost-selected LLM routers violate fixed risk--coverage contracts, including the responsible constraint and violation magnitude. \emph{Second}, we develop a cost-tolerant commitment interface that converts audit evidence into an eligible routing menu and an explicit retain/replace/defer action; we quantify the resulting reliability, deployment, and serving-cost tradeoffs against strong fixed-margin and formal baselines. \emph{Third}, we show how the eligible policy menu and deployment outcomes change with additional audit evidence and establish the same commitment problem with learned semantic routers.

\begin{figure*}[t]
\centering
\includegraphics[width=.96\textwidth]{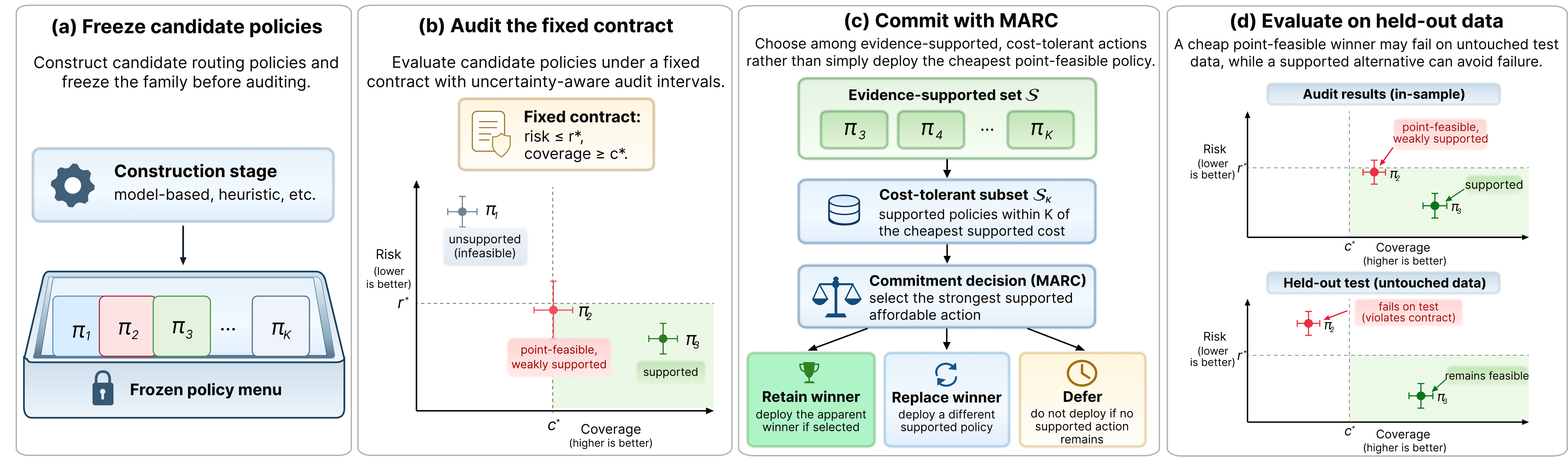}
\caption{MARC overview. Frozen policies are audited against a fixed risk--coverage contract; evidence and cost tolerance determine whether to retain, replace, or defer. Schematic estimates and intervals.}
\label{fig:concept}
\end{figure*}

\section{Related Work}\label{sec:related}
\paragraph{Learning and evaluating economical routers.}
FrugalGPT, Hybrid LLM, RouteLLM, and CARROT study cascades, learned selectors, and prompt-conditioned choices that balance model quality and inference cost \citep{chen2023frugalgpt,ding2024hybrid,ong2024routellm,somerstep2025carrot}. RouterBench, RouterEval, and LLMRouterBench expose heterogeneous model capabilities and costs across tasks \citep{hu2024routerbench,huang2025routereval,li2026llmrouterbench}. MESS+ and SLARouter further study service-constrained, dynamic, and online routing \citep{woisetschlager2025mess,woisetschlager2026slarouter}. MARC complements this policy-construction literature by studying the downstream choice of which frozen routing configuration to authorize from independent audit evidence under a fixed contract.

\paragraph{Selective prediction and calibrated routing.}
Selective classification studies error conditional on answering and its associated coverage \citep{chow1970optimum,elyaniv2010foundations,geifman2017selective}; Conformal Risk Control supplies guarantees for suitable bounded-loss calibration problems \citep{angelopoulos2024conformalrisk}. Routing-specific methods such as RACER, conformal LLM routing, and CP-Router calibrate model sets, gates, or uncertainty-guided model choices \citep{hao2026racer,uddin2026conformal,su2026cprouter}. MARC instead evaluates a complete, frozen policy menu against a joint accepted-risk ceiling and coverage floor, then connects service evidence to affordability and the final policy-level deployment action.

\paragraph{Evidence-supported deployment and multiple testing.}
High-Confidence Policy Improvement established statistical support as a basis for deployment decisions \citep{thomas2015highconfidence}; model-selection bias and adaptive evaluation motivate accounting for choices among candidates \citep{cawley2010overfitting,dwork2015preserving,berk2013valid}. Learn-then-Test and Pareto Testing provide established routes to risk control and optimization under multiple constraints \citep{angelopoulos2022learn,laufer2023pareto}, with Holm, Bentkus-type bounds, and simultaneous resampling supplying relevant statistical tools \citep{holm1979simple,bentkus2004hoeffding,westfall1993resampling,vanderVaartWellner1996}. Recent LLM-routing diagnostics additionally distinguish model-pool opportunity from realizable router performance \citep{shihab2026opportunity}. We place these approaches on a common fixed-contract routing decision problem and measure their consequences for the selected policy, deployment frequency, held-out violations, and serving cost. This makes the commitment decision itself---rather than only an estimated policy score---the object of evaluation.

\section{Fixed-Contract Routing Deployment}\label{sec:problem}
A policy $\pi\in\mathcal P$ maps a query $Z$ to a model action or abstention. Let $A_\pi(Z)\in\{0,1\}$ indicate an answer, and let $\ell_\pi(Z)\in[0,1]$ be the loss on answered requests. Coverage and selective risk are
\begin{equation}\label{eq:contract}
 C(\pi)=\mathbb E[A_\pi],\qquad
 R(\pi)=\frac{\mathbb E[A_\pi\ell_\pi]}{\mathbb E[A_\pi]},
\end{equation}
with $R$ defined at positive coverage. The population service contract is $R(\pi)\le r^\star$ and $C(\pi)\ge c^\star$. We distinguish the policy-level serving-cost objective $J(\pi)$, computed per incoming query under the benchmark's abstention convention, from economic cost of refusing to deploy a policy at all. The latter may involve fallback or latency and is not observed as a single number in the traces.

The numerical pair $(r^\star,c^\star)$ and the finite menu $\mathcal P$ are fixed independently of the commitment audit. Using audit estimates $\widehat R,\widehat C,\widehat J$, the cheapest apparently feasible winner is
\begin{equation}\label{eq:winner}
\begin{aligned}
\widehat{\mathcal M} &= \{\pi:\widehat R(\pi)\le r^\star,\\
 &\hspace{1.7em}\widehat C(\pi)\ge c^\star\},\\
\widehat\pi_{\rm win} &= \arg\min_{\pi\in\widehat{\mathcal M}}\widehat J(\pi).
\end{aligned}
\end{equation}
An empty feasible menu implies deferral. Define the joint population and empirical contract margins
\begin{equation}\label{eq:margin}
\begin{aligned}
m(\pi)&=\min\{r^\star-R(\pi), C(\pi)-c^\star\},\\
\widehat m(\pi)&=\min\{r^\star-\widehat R(\pi),\widehat C(\pi)-c^\star\}.
\end{aligned}
\end{equation}
A point-feasible winner has $\widehat m(\pi)\ge0$, not necessarily $m(\pi)\ge0$. The problem combines two sources of uncertainty: empirical margins for any fixed policy can be wrong, and selection among many audit-evaluated candidates can exploit such fluctuations. The independent-gate comparator fixes a winner on a separate selection sample before applying a point-feasibility test on audit data. Because its deployed population differs, its outcome is not a causal estimate of selection bias removed; it establishes that an additional point gate remains insufficient evidence in this evaluated setting.

The data chronology is construction $\rightarrow$ freeze routing menu and numerical targets $\rightarrow$ independent commitment audit $\rightarrow$ commit $\rightarrow$ untouched held-out test. A committed policy is evaluated on fresh test samples without changing its thresholds. Let $D$ indicate commitment and $F$ a fresh contract violation, $R_{\rm test}>r^\star$ or $C_{\rm test}<c^\star$. We report deployment $P(D)$, unconditional deploy-and-fail $P(D\cap F)$, conditional failure $P(F\mid D)$ (undefined when no deployment), and severity
\begin{equation}\label{eq:severity}
 v=\max\{R_{\rm test}-r^\star,\ c^\star-C_{\rm test},\ 0\}.
\end{equation}
Observed held-out failure is a measurable experiment outcome, whereas an audit's simultaneous-support statement is a statistical guarantee subject to its assumptions. Neither implies the other mechanically. Our primary inferential unit is the dataset, not every seed/contract/cost cell, because the latter share underlying examples and model outcomes.

\section{MARC: Evidence-Aware Commitment}\label{sec:method}
\MARC{} is a commitment interface applied \emph{after} candidate routers have been constructed. It qualifies candidates against a fixed contract, forms an affordable subset of the qualified menu, and deploys a policy with relatively strong evidence or defers. The two versions below differ in evidence construction, not in the deployment decision.

\paragraph{MARC-P: policywise practical evidence.}
For policy $\pi$, bootstrap the shared query-level audit records to form one-sided bounds $U_R^{\alpha}(\pi)$ for accepted-query risk and $L_C^{\alpha}(\pi)$ for coverage. The eligible menu is
\begin{equation}\label{eq:pointset}
\begin{aligned}
\mathcal B_{\rm P}=\{\pi:\; &r^\star-U_R^{\alpha}(\pi)\ge \delta_R,\\
 &L_C^{\alpha}(\pi)-c^\star\ge\delta_C\}.
\end{aligned}
\end{equation}
Explicit buffers $\delta_R,\delta_C$ are fixed; bootstrap adjustments vary with the audit evidence and policy. \emph{Policywise} intervals do not, by themselves, validate arbitrary post-selection choice over the menu. Accordingly, MARC-P's small test failure rates are empirical observations, not an unconditional finite-sample familywise certificate.

\paragraph{MARC-S: menu-level simultaneous evidence.}
For policy $\pi$, let $\widehat m^{(b)}(\pi)$ be its margin on common query-level bootstrap resample $b$. Set
\begin{equation}\label{eq:simstat}
\begin{aligned}
T^{(b)}&=\max_{\pi\in\mathcal P}\{\widehat m(\pi)-\widehat m^{(b)}(\pi)\},\\
q_{1-\alpha}&=Q_{1-\alpha}\{T^{(b)}\}_{b=1}^{B}.
\end{aligned}
\end{equation}
The common correction yields
\begin{equation}\label{eq:simset}
 \mathcal B_{\rm S}(\delta)=\{\pi:\widehat m(\pi)-q_{1-\alpha}\ge\delta\}.
\end{equation}
Shared resamples preserve policy dependence; valid simultaneous coverage extends to an audit-selected policy from the frozen menu. Appendix~\ref{app:validity} gives the conditional finite-menu asymptotic result and a proof, not a distribution-free finite-sample guarantee. Positive population coverage, separated active margin components, and bootstrap regularity matter, as do the source's finite-$B$ upper empirical-quantile semantics.

\paragraph{Affordable selection and three actions.}
For either eligible menu $\mathcal B$, defer if it is empty. Otherwise select among candidates within a frozen relative cost tolerance of the least expensive eligible policy:
\begin{equation}\label{eq:kappa}
\mathcal B_\kappa=\{\pi\in\mathcal B:\widehat J(\pi)\le(1+\kappa)\min_{\rho\in\mathcal B}\widehat J(\rho)\}.
\end{equation}
MARC-P ranks this affordable menu by the sum of risk and coverage \emph{bound-adjusted} slacks; MARC-S ranks by the simultaneous lower joint margin. Deterministic ties use cost and frozen policy ID. The output is \emph{retain} if the selected policy equals the cheapest apparent audit winner, \emph{replace} if it is different, and \emph{defer} if no qualified policy exists. The output therefore changes the routing policy or withholds deployment, rather than simply testing the cheapest candidate.

The displayed MARC settings are nominal $\alpha=.10$, $\kappa=.25$, $B=1000$, and explicit $\delta_R=\delta_C=.05$ for MARC-P; the frozen MARC-S menu contains $\delta\in\{0,.02,.05\}$. These rules have different validity scopes; implementation details and tie conventions appear in Appendix~\ref{app:algorithms}.

\section{Experimental Design}\label{sec:setup}
\paragraph{Primary RouterBench study.}
We use frozen cached model quality and cost outcomes from RouterBench \citep{hu2024routerbench}, covering 19 task families and deterministic menus of simple/selective policies. For each of 20 split seeds, 20\% of the data constructs the policies, 20\% provides the independent deployment audit, and 60\% is untouched test data. The fixed contracts are all six combinations of $r^\star\in\{.30,.50,.70\}$ and $c^\star\in\{.40,.60\}$; the evaluation includes the frozen binary and bounded correctness channels and separate mean and 90th-percentile tail serving-cost objectives. MARC uses $B=1000$ bootstrap replicates; full provenance appears in Appendix~\ref{app:protocol}.

\paragraph{Learned semantic-routing study.}
A second fixed-contract experiment uses a subset of LLMRouterBench \citep{li2026llmrouterbench}: 10 datasets, 12,444 queries, 11 models and a menu of 48 CARROT-style KNN operating policies \citep{somerstep2025carrot}. Query embeddings inform estimated model suitability; frozen cost preferences and acceptance thresholds define the menu. The split is 40\% fitting, 20\% policy selection, 20\% independent audit and 20\% untouched test, repeated for 20 seeds. The experiment covers one learned architecture on a selected ten-dataset subset (Appendix~\ref{app:learned}).

\paragraph{Comparator and evaluation scope.}
We compare the cheapest point-feasible winner; an independently selected winner subsequently audited with a point gate; the full frozen invariant-margin family ($\delta\in\{0,.02,.05,.08,.10,.15\}$); MARC-P/S; Hoeffding--Bentkus LTT with Bonferroni and Holm multiplicity; Pareto Testing; and exact-binomial Holm/Bonferroni tests where the score truly is Bernoulli \citep{angelopoulos2022learn,laufer2023pareto}. Nominal $\alpha=.10$ comparisons use applicable methods on compatible score domains; formal-only $\alpha=.05$ and binary exact controls are reported separately. The stricter $(r^\star,c^\star)=(.30,.60)$ contract is an illustrative member of the original grid, not an endpoint nominated after seeing its outcome.

\paragraph{Sampling and outcome estimands.}
Each dataset's repeated seeds, contracts, score settings, and cost objectives are summarized first and dataset means are then weighted equally in primary RouterBench tables. The 19 task families include 12 MMLU-derived subdomains, so these are reporting clusters rather than 19 distinct benchmark sources. We supply dataset-cluster bootstrap intervals and leave-one-dataset-out sensitivity rather than counting 9,120 correlated decision records per method as independent statistical replicates. Learned-router percentages are pooled; conditional metrics are undefined when their denominators are zero. Serving cost is per incoming query with zero policy-level abstention cost, reported among commitments; fallback and commitment-stage deferral are unpriced.

\paragraph{Evidence-size sensitivity.}
The separate, pre-frozen audit-size study uses a 20\% construction sample, a 40\% audit reservoir, and a 40\% common untouched test. Deterministic query hashes create nested $A_{10}\subset A_{20}\subset A_{40}$ audits of 10\%, 20\%, and 40\% of each full dataset. Policies, contracts, settings, and test data remain fixed; MARC-P is compared with the complete fixed-margin sweep.

\section{Results}\label{sec:results}
\subsection{Why independent point feasibility can fail}\label{sec:resultfailure}
The cheapest audit-feasible policy deploys in 91.82\% of RouterBench opportunities, yet 32.22\% of those deployments violate the fresh contract (29.63\% unconditional deploy-and-fail; Table~\ref{tab:primary}). An independent point-feasibility gate, applied to a candidate selected on separate data, deploys in only 63.00\% of cases and still fails in 21.08\% of its deployments. Thus an extra validation split reduces exposure but does not turn point feasibility into evidence-backed authorization. The comparison highlights the practical difference between independently checking point feasibility and accounting for uncertainty when authorizing a policy: a separate gate still leaves substantial held-out violations.

The winner's failures are not merely infinitesimal threshold crossings: among violating deployments, the maximum risk/coverage excess averages 3.15 percentage points (median 1.94, p90 7.39). In the full grid, 10.83\% of opportunities incur risk-only violations, 16.64\% coverage-only, and 2.15\% both (Appendix~\ref{app:failure}). For MARC-P $.05$, the corresponding unconditional violation rates are 0.54\% risk-only, 0.09\% coverage-only, and 0.00\% joint, compared with 10.83\%, 16.64\%, and 2.15\% for the audit winner. Coverage misses are especially consequential for selective routing: a policy can improve accepted-query accuracy by abstaining more often yet breach its minimum service-availability commitment. These are observed failure compositions of different deployment decisions, not effects of an isolated algorithm component.

The frozen strict contract $(r^\star,c^\star)=(.30,.60)$ exhibits the same problem: the winner deploys in 81.32\% of cells (652 failures/1,236 deployments), versus MARC-P's 65.00\% (8/988). MARC-S $.02$ and LTT-Holm have no observed failures among 600 and 772 deployments, respectively. These are finite held-out counts, not guarantees of zero population failure. The six-contract grid remains the primary evaluation; the strict cell is illustrative.

\begin{table}[t]
\centering\small\setlength{\tabcolsep}{3pt}
\begin{tabular}{lrrr}\toprule
Procedure & Deploy & D\&F & Fail$\mid$D \\
\midrule
Audit winner & 91.82 & 29.63 & 32.22\\
Independent gate & 63.00 & 12.38 & 21.08\\
Point margin $.00$ & 91.82 & 22.25 & 24.40\\
Fixed margin $.05$ & 87.72 & 6.06 & 7.25\\
Fixed margin $.10$ & 86.32 & 1.43 & 1.78\\
Fixed margin $.15$ & 82.74 & 0.29 & 0.38\\
\MARCP{} $.05$ & 85.79 & 0.63 & 0.73\\
Pareto Testing & 84.41 & 0.49 & 0.60\\
LTT-Holm & 78.38 & 0.09 & 0.12\\
\MARCS{} $.02$ & 71.93 & 0.13 & 0.21\\
\MARCS{} $.05$ & 68.03 & 0.00 & 0 observed\\
\bottomrule\end{tabular}
\caption{RouterBench full-grid results (dataset-equal percentages). D\&F: deploy-and-fail; Fail$\mid$D: dataset-averaged conditional failure, not necessarily D\&F/Deploy. MARC suffixes denote buffers $\delta$; guarantee scopes differ. Binary-only exact tests appear in Appendix~\ref{app:exact}.}\label{tab:primary}
\end{table}

\subsection{What evidence changes in the deployment decision}\label{sec:frontier}
The operating points reveal how both feasibility evidence and policy ranking shape deployment. The cheapest winner deploys in 91.82\% of opportunities with 29.63\% deploy-and-fail. Zero-margin, cost-tolerant reranking maintains deployment while lowering failure exposure to 22.25\%, but its effect depends strongly on the objective: only $29.52\%\to29.17\%$ for mean cost versus $29.74\%\to15.33\%$ for tail cost (Table~\ref{tab:v3_rerank}). A fixed $.05$ margin yields 87.72\% deployment and 6.06\% deploy-and-fail; MARC-P $.05$ yields 85.79\% and 0.63\%. Reranking matters, but cannot substitute for supported feasibility. The mean-cost result is especially instructive: changing the selected point-feasible policy alone barely changes the failure exposure, whereas adding margin or uncertainty evidence sharply reduces it. For tail cost, changing the ranking itself can have a large effect, so the contributions of eligibility and ranking are not interchangeable across economic objectives.

\begin{table}[!htbp]\centering\small\setlength{\tabcolsep}{4pt}
\begin{tabular}{lrrrr}\toprule
Objective & Winner & Fixed $.00$ & $\Delta$ & MARC-P $.05$ \\
\midrule
Mean cost & 29.52 & 29.17 & -0.35 & 0.77 \\
Tail cost & 29.74 & 15.33 & -14.41 & 0.48 \\
\bottomrule\end{tabular}
\caption{Deploy-and-fail by optimization objective (percent). $\Delta$ is fixed $.00$ minus the audit winner, in percentage points.}\label{tab:v3_rerank}\end{table}

The action comparison (Table~\ref{tab:actions}) makes the deployment decision tangible. MARC-P $.05$ replaces the apparent winner in 77.11\% of opportunities and defers in 14.21\%; its improvement therefore involves authorizing different routing policies, not solely withholding service. Pareto Testing and LTT-Holm also replace many winners, but the methods produce different retain/replace/defer mixes, with corresponding reliability, availability, and serving-cost consequences.

\begin{table}[t]
\centering
\footnotesize
\setlength{\tabcolsep}{3.3pt}
\begin{tabular}{lrrr}
\toprule
Procedure & Retain & Replace & Defer\\
\midrule
Audit winner & 91.82 & 0.00 & 8.18\\
LTT-Holm & 8.64 & 69.74 & 21.62\\
MARC-P .05 & 8.68 & 77.11 & 14.21\\
MARC-S .02 & 8.46 & 63.46 & 28.07\\
Pareto & 17.81 & 66.60 & 15.59\\
\bottomrule
\end{tabular}
\caption{RouterBench commitment actions (percent of all decision opportunities; dataset-equal).}
\label{tab:actions}
\end{table}

More stringent evidence rules occupy different operating points. MARC-P has 85.79\% deployment and 0.73\% conditional failure; Pareto Testing has 84.41\% and 0.60\%, while LTT-Holm has 78.38\% and 0.12\%. MARC-S at $.02$ deploys in 71.93\% with 0.21\% observed conditional failure. Conservative fixed margins are also competitive. The procedures offer distinct combinations of observed reliability and deployment, with statistical guarantees specific to each construction. MARC-P uses practical policywise bounds; MARC-S uses a simultaneous asymptotic correction; and formal tests supply their corresponding risk-control guarantees. These differences make serving cost and the frequency of withheld commitments essential to comparing operating points.

\begin{figure*}[t]
\centering
\includegraphics[width=.96\textwidth]{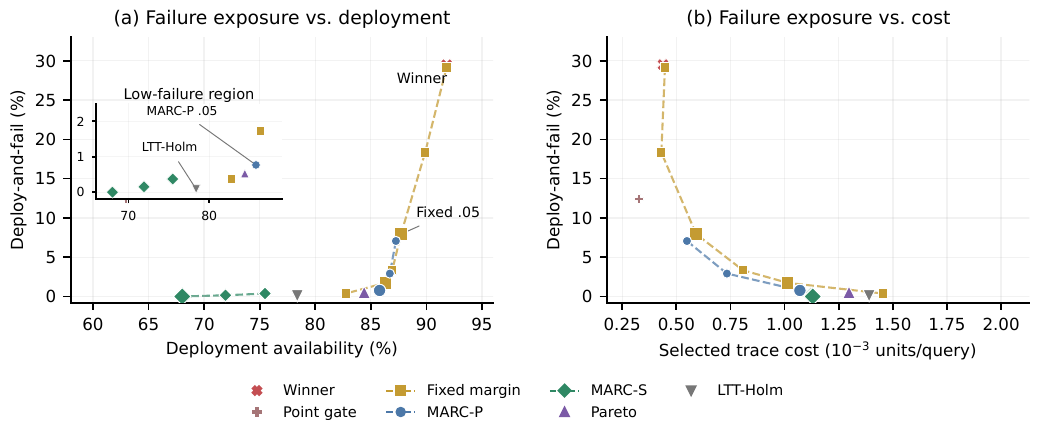}
\caption{Mean-cost RouterBench operating points: (a) deployment versus deploy-and-fail; (b) selected serving cost versus deploy-and-fail. Costs are trace units per incoming query among deployed policies. Dashed lines connect frozen parameter sweeps, not interpolated policies; deferral costs are unmeasured.}\label{fig:frontier}
\end{figure*}

These reliability gains have a serving-cost consequence (Figure~\ref{fig:frontier}). Under mean-cost optimization, the deployed policy costs about $0.000438$ trace units per incoming request for the winner versus $0.001071$ for MARC-P, using each rule's committed population. Matching only opportunities where both deploy gives a MARC-P premium of $0.000844$ over the winner (3,912 paired decisions; 95\% dataset-bootstrap interval $[0.000408,0.001444]$). Against fixed margin $.10$, the matched difference is $+0.000070$ (interval $[-0.000019,0.000159]$); the interval crosses zero. Comparisons with Pareto and LTT-Holm instead yield $-0.000243$ and $-0.000517$ on their respective shared deployment populations. These pairwise comparisons have different matching sets and exclude fallback or nonservice costs. Mean and tail serving-cost objectives are also distinct; Appendix~\ref{app:economics} reports objective-specific estimates, uncertainty, and frozen penalty scenarios. The method-specific deployed averages answer what a rule pays on the opportunities it actually commits; common-deployment comparisons instead isolate the cost of the replacement policy on identical opportunities. Neither estimates the economic effect of cases where only one procedure deploys. This distinction matters because an apparent cost reduction may arise from serving a different mix of tasks rather than routing the same tasks more economically.

\begin{figure*}[t]
\centering
\includegraphics[width=.96\textwidth]{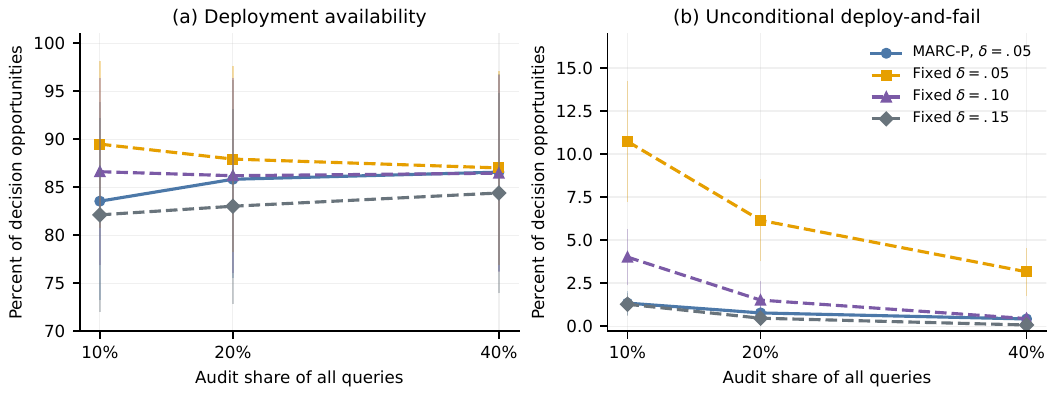}
\caption{Nested-audit comparison at 10\%, 20\%, and 40\%: (a) deployment and (b) deploy-and-fail, with between-dataset intervals. MARC-P and the fixed-margin controls keep their settings unchanged. Paired audit-size effects appear in Appendix~\ref{app:audit}.}\label{fig:audit}
\end{figure*}

\subsection{When additional audit information becomes available}\label{sec:auditsize}
The nested-audit study holds policies, fixed contracts, method settings, and a common test set constant while expanding the audit $A_{10}\subset A_{20}\subset A_{40}$. It uses a separate 20/40/40 construction/audit-reservoir/test partition and paired comparisons within dataset and seed (Appendix~\ref{app:audit}).

As the audit grows from 10\% to 40\%, MARC-P $.05$ increases deployment from 83.55\% to 86.58\% while reducing deploy-and-fail from 1.34\% to 0.42\% (Figure~\ref{fig:audit}). The paired A40--A10 differences are $+3.03$ and $-0.92$ percentage points; mean eligible-set size rises by about 1.49 policies. Dataset-bootstrap intervals are $[+1.01,+5.44]$ and $[-1.67,-0.22]$ points, respectively. The increase in eligible-set size shows how additional audit evidence can make more cost-aware commitments available. The response varies by dataset: nine gain deployment, two lose it, and eight are unchanged between A10 and A40. Paired shifts and supported-menu expansion together characterize the practical effect of collecting more audit data.

On opportunities deployed at both audit sizes, MARC-P also selects lower-cost policies at A40 than at A10 ($\Delta J=-0.000193$ in frozen trace units per incoming query; Appendix~\ref{app:audit}). This matched comparison isolates a change in policy choice, while the deployment-rate increase captures newly authorized opportunities. The two measures together help distinguish benefits from a larger eligible menu from benefits that merely reflect serving fewer requests.

Fixed margins offer informative alternatives. The $.05$ buffer has 89.50\% deployment and 10.73\% deploy-and-fail at A10, versus 87.02\% and 3.16\% at A40; fixed $.15$ is comparably conservative to MARC-P at A10. At A40, fixed $.10$ closely matches MARC-P (86.47\% versus 86.58\% deployment; 0.44\% versus 0.42\% deploy-and-fail). MARC-P's policy-specific uncertainty adjustment responds to additional evidence without changing its preset numerical buffer, offering a practical way to manage commitments across audit-size regimes while preserving competitive fixed-margin choices.

\subsection{The same deployment issue arises in learned semantic routing}\label{sec:learned}
In the learned CARROT-style study, query embeddings inform model-suitability predictions, while 48 frozen policies vary cost preferences and acceptance thresholds. The learned-menu protocol separates fitting, operating-policy selection, commitment auditing, and final testing. Unlike the hand-designed policy families, these routers make text-dependent model choices; the same fixed-contract commitment issue persists when routing policies are learned before the audit.

\begin{table*}[t]
\centering\footnotesize\setlength{\tabcolsep}{3.6pt}
\begin{tabular}{lrrrr|rrr}\toprule
&\multicolumn{4}{c}{Full frozen grid ($N=1{,}200$)}&\multicolumn{3}{c}{Illustrative strict contract ($N=200$)}\\
\cmidrule(lr){2-5}\cmidrule(lr){6-8}
Procedure & Deploy & D\&F & Fail$\mid$D & Fail/dep. & Deploy & Fail$\mid$D & Fail/dep.\\
\midrule
Audit winner &81.92&27.83&33.98&334/983&67.50&41.48&56/135\\
Fixed $.05$&76.83&9.42&12.26&113/922&62.50&20.00&25/125\\
\MARCP{} $.05$&69.00&2.58&3.74&31/828&45.50&8.79&8/91\\
Pareto Testing&58.25&0.50&0.86&6/699&25.50&0 obs.&0/51\\
LTT-Holm&51.33&0.17&0.32&2/616&21.00&0 obs.&0/42\\
\MARCS{} $.02$&56.17&0.50&0.89&6/674&23.00&0 obs.&0/46\\
\MARCS{} $.05$&53.17&0.25&0.47&3/638&20.00&0 obs.&0/40\\
\bottomrule\end{tabular}
\caption{Learned-router commitments (pooled percentages; mean-cost objective). Full grid: $N=1{,}200$; strict $(r^\star,c^\star)=(.30,.60)$: $N=200$. Fail/dep.: observed failure/deployment counts.}\label{tab:learned}
\end{table*}

In the pooled learned-router full grid (Table~\ref{tab:learned}), the winner fails on 334 of 983 commitments (33.98\%) at 81.92\% deployment. MARC-P $.05$ fails on 31 of 828 (3.74\%) at 69.00\% deployment; MARC-S $.02$, Pareto Testing, and LTT-Holm achieve lower observed failure with fewer deployments. Unlike dataset-equal RouterBench aggregates, these results pool 1,200 learned-router decision opportunities. The strict contract in the same table isolates a more demanding service target without changing the frozen evaluation protocol.

At the strict $(.30,.60)$ contract, MARC-P deploys in 45.50\% of opportunities with 8 failures among 91 commitments. MARC-S $.02$, Pareto, and LTT-Holm observe no failures among 46, 51, and 42 commitments but authorize only 23.00\%, 25.50\%, and 21.00\% of opportunities, respectively. The strict-contract contrast makes the availability cost of stronger evidence requirements visible, alongside the reduction in observed failures.

The pooled average also masks distinct task-level decisions. On ArenaHard, MARC-P authorizes 108 of 120 opportunities with no observed failures, compared with 39 failures among 120 winner deployments. On LiveCodeBench, it authorizes all 120 opportunities with two observed failures, versus 41 for the winner. On \texttt{tau2}, it authorizes 83 of 120 with six failures, compared with 58 failures in the winner's 114 deployments. HLE and SWE-bench present the opposite operational outcome: MARC-P defers every opportunity in this panel. These contrasts (Appendix~\ref{app:learned}) show why both authorization rate and conditional failure matter when a frozen learned-router menu is evaluated across heterogeneous NLP tasks.

Serving cost follows the same practical tradeoff. Among deployed learned-router policies under mean-cost optimization, the winner costs $0.007221$ trace units per incoming query and MARC-P $.05$ costs $0.012464$; Pareto and LTT-Holm cost $0.012916$ and $0.012046$, respectively. These method-specific committed populations differ, so the values describe selected serving cost rather than a matched end-to-end cost comparison. Residual failures are rarer under MARC-P but are not uniformly mild: conditional mean severity is $0.0709$, versus $0.0624$ for the winner. Detailed per-task counts and severity appear in Appendix~\ref{app:learned}.

\section{Discussion and Conclusion}\label{sec:discussion}
Cost-aware router evaluation requires a deployment decision beyond choosing the cheapest policy that passes empirical thresholds. Across frozen policies and learned semantic routing, independent audit estimates can favor configurations with insufficient service-contract headroom. MARC makes that last decision explicit: evidence-supported eligibility and serving-cost tolerance determine whether the apparent winner is retained, replaced by another router, or deferred.

These choices have measurable consequences. MARC-P typically replaces the cheapest apparent winner rather than only withholding deployments; formal controls and fixed margins offer different combinations of reliability, deployment rate, and serving cost. As audit evidence grows, MARC-P supports more policies and can authorize more deployments with fewer observed violations. These results expose a practical reliability--deployment--cost tradeoff. Evaluating a commitment requires accounting for the reliability of its service claims, the policies it authorizes, the opportunities it defers, and the serving cost of those choices.

\section*{Limitations}
The results come from cached benchmark outcomes and randomized independent partitions, not live traffic subject to domain shift, model updates, incomplete feedback, or changing service promises. The learned semantic panel covers ten datasets, eleven models and one KNN router family, with sparse deployment on some tasks. The 19 RouterBench task families, not repeated cells within them, are the reporting clusters for descriptive benchmark uncertainty. Because 12 are MMLU-derived subdomains, cluster bootstrap intervals describe variation across the evaluated task panel rather than independent generalization across 19 distinct benchmark sources. MARC-P's policywise bootstrap is not a generic finite-sample post-selection certificate, and MARC-S's joint-menu bootstrap justification is conditional and asymptotic under positive coverage and active-constraint regularity. Formal controls have their own null, score-channel and familywise assumptions. The evaluation measures specific frozen menus and cannot assign a monetary value to withheld deployment, fallback, delays, or service-contract violations. Contract thresholds are fixed before auditing.

\section*{Ethical Considerations}
The study analyzes cached evaluation traces and does not deploy policies to users. Low average selective loss or a formal statistical threshold does not establish suitability for high-stakes decisions, nor does it ensure fairness across user subpopulations. Production deployments require independently justified losses, coverage requirements, monitoring, safe fallback behavior, and re-auditing under distribution shift.

\bibliographystyle{plainnat}
\bibliography{ref}
\appendix
% Appendix-only float policy: favor local placement without hard pinning.
\renewcommand{\topfraction}{.90}
\renewcommand{\bottomfraction}{.80}
\renewcommand{\textfraction}{.08}
\renewcommand{\floatpagefraction}{.60}
\renewcommand{\dbltopfraction}{.88}
\renewcommand{\dblfloatpagefraction}{.58}
\setcounter{topnumber}{4}
\setcounter{bottomnumber}{3}
\setcounter{totalnumber}{7}
\setcounter{dbltopnumber}{4}

\section{Data, Contracts, and Experimental Chronology}\label{app:protocol}
\paragraph{Data scope and inferential unit.} RouterBench's 19 dataset/task families are the reporting clusters used for descriptive benchmark-level uncertainty. In the frozen full-grid experiment, each method has 9,120 decision records: 19 datasets, 20 seeds, six risk--coverage contracts, two cost objectives, and applicable score channels; the binary-only exact-testing panel has a smaller domain. Repetitions from one dataset share cached LLM outcomes and are not treated as 9,120 independent population samples. The learned panel has ten task datasets and its mean-cost comparison has 1,200 frozen decision opportunities over all contracts and seeds. Its pooled numbers are intentionally not pooled with the dataset-equal RouterBench metrics.
\paragraph{Trace costs and abstentions.} Model inference cost is inherited from the frozen benchmark trace and assembled per routing policy. A policy that abstains on a request receives the frozen zero policy-level serving-cost convention on that request. The resulting mean-cost objective is therefore per \emph{incoming} query, not per answered request. A tail-cost objective separately summarizes an upper-quantile cost functional. The commitment-stage act of declining to deploy a policy is a different event; its downstream fallback or nonservice cost was not measured. All economic results are conditional on deployment decisions unless stated otherwise.
\paragraph{Zero accepted queries.} A policy accepting no audit queries has undefined conditional-risk estimation and cannot be treated as satisfying the risk constraint. Positive population coverage is separately required for MARC-S's asymptotic result. For ARC Challenge, inference costs use the per-query token/length aggregation in the evaluation trace.

\paragraph{Evaluation units.} A decision opportunity consists of a dataset, split seed, frozen candidate menu, loss channel, cost objective, and fixed risk/coverage contract. The test sample is used solely to assess the selected policy. RouterBench has 19 task families and 20 seeds. Its primary construction/audit/test shares are 20/20/60 percent. LLMRouterBench has 10 tasks and a 40/20/20/20 fit/selection/audit/test chronology, 20 seeds, and 48 deterministic KNN policies. Contract thresholds are enumerated before the audit; no downstream target selection is permitted. Repeated split seeds and contract cells are not independent task replications.

\paragraph{Model and candidate-policy scope.} The primary menu uses the frozen policies from the RouterBench input manifest: each construction split selects a least-cost model and a strongest model using construction-only statistics, then defines preset selective/length- and confidence-threshold operating policies. The frozen construction uses three prompt-length quantiles ($.25,.50,.75$), three confidence thresholds ($.55,.65,.75$), and four acceptance thresholds ($.40,.55,.70,.85$), alongside predefined policy IDs and tie ordering. These policies are evaluated as an \emph{already constructed} menu, not tuned on audit or test outcomes. The learned panel instead fixes 48 CARROT-style KNN operating policies using independent fit and selection splits. The accompanying reproducibility package provides the evaluated model pools, filtering rules, construction code, and source manifests.

\paragraph{Frozen settings.} RouterBench risk targets are $r^\star\in\{.30,.50,.70\}$ and minimum coverage targets $c^\star\in\{.40,.60\}$. Both bounded correctness and binary correctness channels are included where valid, and the mean-cost and 90th-percentile tail-cost objectives are kept separate for economic inference. Main MARC experiments use $B=1000$, nominal $\alpha=.10$, and $\kappa=.25$; MARC-P's principal explicit risk and coverage margins are $.05$ each and the MARC-S simultaneous buffer is varied on its frozen grid. Bounds and risk estimators are recomputed from accepted-query observations; a zero-acceptance bootstrap replicate is handled conservatively under the frozen canonical rule. Identical resampled query indices are applied to all frozen policies within a replicate.

\paragraph{Construction and reproducibility.} Two independent executions of the RouterBench analysis with $B=1000$ yielded identical 200,640 commitment decisions and 319,200 intermediate audit records. For learned routing, deterministic reconstruction from verified traces and embeddings matched 800 reference router diagnostics and 30,438 policy-decision metrics; repeated evaluations agreed. Reported results use these validated records.

\paragraph{Sample size and validity.} Formal testing baselines use their canonical Hoeffding--Bentkus p-values and multiple-testing conventions. Exact-binomial controls use Bernoulli outcomes and therefore are not applied to fractional losses. The same nominal $\alpha$ need not entail the same theoretical guarantee across MARC-P, MARC-S, and formal controls. Zero observed failures indicate a sample result, not a zero population violation probability.

\section{Commitment Procedures and Formal Comparators}\label{app:algorithms}
\paragraph{Commitment-rule pseudocode.}
\begin{enumerate}[leftmargin=*,itemsep=1pt,topsep=2pt]
\item \textbf{Freeze inputs:} candidate menu, $(r^\star,c^\star)$, policy IDs, loss channel, cost objective, method, $\alpha,\delta,\kappa$, and bootstrap seed; none may depend on commitment-audit outcomes.
\item \textbf{Audit:} calculate each policy's accepted-query risk, coverage, objective-specific serving cost, and any method-specific bootstrap bounds or valid component p-values. A policy with zero accepted audit queries has undefined/infinite risk and cannot pass the risk requirement.
\item \textbf{Qualify:} construct the method-specific set $\mathcal B$ using point margins (fixed-buffer control), policywise bootstrap inequalities (MARC-P), simultaneous lower joint margins (MARC-S), or the formal testing/ordered-frontier acceptance rule. The empirical audit winner is selected by minimum feasible cost instead.
\item \textbf{Commit:} if $\mathcal B=\varnothing$, defer; otherwise restrict to $(1+\kappa)$ times the cheapest supported audit cost (zero minimum permits only zero-cost policies). Choose maximum method-specific safety score, then apply the prespecified deterministic cost/policy-ID tie rules. Pareto Testing retains its own cheapest accepted-prefix selection rule.
\item \textbf{Evaluate:} compare the emitted policy with the audit-feasible cheapest policy (retain versus replace); compute held-out contract failure only after the decision is fixed. Record the entire menu and support set, not only the winner.
\end{enumerate}
The methods use distinct margin scores: MARC-P maximizes the sum of its \emph{bound-adjusted} risk and coverage slacks; MARC-S maximizes a simultaneous lower \emph{joint} margin; fixed-margin and LTT controls rank by empirical joint margin within their own eligible and affordable sets. A joint unsafe policy null is handled by the maximum of component risk/noncoverage p-values before applying Holm or Bonferroni across policies. Pareto's ordering arises on its separate optimization split and stops testing at the first policy failing the strict $p<\alpha$ criterion. The final action space is shared, while eligibility tests and selection rules differ.

\paragraph{MARC-P.} The frozen source evaluates one-sided policy-specific bootstrap quantiles of risk and coverage, shifts each point-estimated contract boundary by the resulting uncertainty bound, and enforces explicit buffers $\delta_R,\delta_C$. For MARC-P, the risk-upper quantile uses finite bootstrap replicate risks (and is $+\infty$ if none are finite), while the coverage-lower quantile includes replicates with zero accepted requests; the frozen risk and coverage marginal quantiles use the linear empirical convention. This differs from MARC-S, which maps zero-acceptance replicate joint margins to $-\infty$ and uses the \texttt{higher} convention for its menu-level upper maximum-deviation quantile. If there is no supported policy, MARC-P defers; otherwise the cost-tolerant subset of Eq.~\ref{eq:kappa} is evaluated using the sum of bound-adjusted slacks. The replicate budget and empirical quantile convention are fixed implementation choices.

\paragraph{MARC-S.} The source uses one common maximum downward bootstrap deviation across the \emph{entire frozen policy family}; its empirical upper order statistic (\texttt{higher} convention) is subtracted from every policy's empirical joint margin. This preserves common-resample policy dependence. Cost-tolerant selection then maximizes the resulting lower joint margin. Neither method retrains or changes policy predictions during auditing.

\paragraph{Independent point gate and fixed margins.} The independent gate first constructs a winner on separate selection data, then uses point-estimated risk and coverage from audit data to decide whether to deploy that frozen winner. It is not an uncertainty-corrected feasibility certificate. The fixed-margin family replaces point-feasibility inequalities with $\widehat R\le r^\star-\delta$ and $\widehat C\ge c^\star+\delta$, followed by the same frozen cost tolerance and selection convention where applicable. All $.00,.02,.05,.08,.10,.15$ buffers are retained in the frozen grid.

\paragraph{LTT, Holm/Bonferroni, and Pareto.} Learn-then-Test (LTT) and Pareto Testing are substantial alternatives with different validity premises from MARC-P. The canonical controls apply Hoeffding--Bentkus bounded-loss hypothesis tests \citep{bentkus2004hoeffding}; multiplicity is handled by the named Bonferroni or Holm rule \citep{holm1979simple}, and Pareto uses its canonical fixed-sequence stopping convention. Their guarantees depend on the chosen testing families, bounded-loss assumptions, and calibration protocols; observed test failures are reported separately from those guarantees. Exact-binomial controls employ valid Bernoulli testing only on the binary score domain. The primary comparison uses nominal $\alpha=.10$; the existing formal-only $.05$ rows are reported as sensitivity, never mistaken for matched MARC $.05$ configurations.

\section{Statistical Validity and Its Scope}
\label{app:validity}

This section gives an asymptotic justification for the menu-level
max-deviation correction when the candidate policies and their parameters are
fixed independently of the sample used for the commitment audit.

Let $Z_1,\ldots,Z_n$ denote i.i.d.\ audit observations. For a fixed policy
$\pi$, let $A_\pi(Z)\in\{0,1\}$ indicate acceptance and let
$\ell_\pi(Z)\in[0,1]$ denote loss. Define
\begin{equation}
\begin{aligned}
C(\pi)
&=
\mathbb{E}[A_\pi(Z)],
\\
N(\pi)
&=
\mathbb{E}[A_\pi(Z)\ell_\pi(Z)],
\\
R(\pi)
&=
\frac{N(\pi)}{C(\pi)}.
\end{aligned}
\end{equation}
Their empirical counterparts are denoted by
$\widehat C_n(\pi)$, $\widehat N_n(\pi)$, and
$\widehat R_n(\pi)$. The population and empirical joint margins are
\begin{equation}
m(\pi)
=
\min
\left\{
r^\star-R(\pi),
C(\pi)-c^\star
\right\},
\end{equation}
and
\begin{equation}
\widehat m_n(\pi)
=
\min
\left\{
r^\star-\widehat R_n(\pi),
\widehat C_n(\pi)-c^\star
\right\}.
\end{equation}

Let $\widehat m_n^\ast(\pi)$ denote the corresponding nonparametric-bootstrap
quantity, and define
\begin{equation}
D_n^\ast
=
\max_{\pi\in\mathcal P}
\left\{
\widehat m_n(\pi)
-
\widehat m_n^\ast(\pi)
\right\}.
\end{equation}
Let $q_{1-\alpha,n}^{\mathrm{menu}}$ be the ideal conditional
$(1-\alpha)$ bootstrap quantile of $D_n^\ast$, or a Monte Carlo estimate whose
number of replicates tends to infinity.

\begin{theorem}[Fixed-menu simultaneous validity]
\label{thm:marcs_fixed_menu}
Suppose that:
\begin{enumerate}
    \item
    $\mathcal P$ is finite and fixed independently of
    $Z_1,\ldots,Z_n$;

    \item
    the observations are i.i.d., losses are bounded, and
    \begin{equation}
    \inf_{\pi\in\mathcal P}
    C(\pi)
    \ge
    c_0
    >
    0;
    \end{equation}

    \item
    for some $\eta>0$, every policy has a locally unique active contract
    component:
    \begin{equation}
    \left|
    \bigl\{r^\star-R(\pi)\bigr\}
    -
    \bigl\{C(\pi)-c^\star\bigr\}
    \right|
    \ge
    \eta;
    \end{equation}

    \item
    the limiting distribution of the menu-level maximum is continuous at its
    $(1-\alpha)$ quantile.
\end{enumerate}
Then
\begin{equation}
\begin{aligned}
\liminf_{n\rightarrow\infty}
\Pr\Bigl\{
&
m(\pi)
\ge
\widehat m_n(\pi)
-
q_{1-\alpha,n}^{\mathrm{menu}}
\\
&
\text{for every }
\pi\in\mathcal P
\Bigr\}
\ge
1-\alpha.
\end{aligned}
\end{equation}
\end{theorem}

\begin{proof}
For a finite menu, the vector containing
$\widehat N_n(\pi)$ and $\widehat C_n(\pi)$ for all
$\pi\in\mathcal P$ satisfies a multivariate central limit theorem, and the
ordinary nonparametric bootstrap consistently estimates its joint limiting
distribution. Because coverage is bounded away from zero, the map
\[
(N,C)
\longmapsto
\frac{N}{C}
\]
is smooth in a neighborhood of the population values. The active-component
separation condition makes the minimum defining $m(\pi)$ locally equal to one
smooth branch.

Write
$\widehat{\boldsymbol m}_n
=
(\widehat m_n(\pi))_{\pi\in\mathcal P}$
and
$\boldsymbol m
=
(m(\pi))_{\pi\in\mathcal P}$.

The multivariate bootstrap delta method therefore yields, jointly over the
finite menu,
\begin{equation}
\sqrt n
\left(
\widehat{\boldsymbol m}_n
-
\boldsymbol m
\right)
\rightsquigarrow
\boldsymbol G,
\end{equation}
and, conditionally on the audit sample,
\begin{equation}
\sqrt n
\left(
\widehat{\boldsymbol m}_n^\ast
-
\widehat{\boldsymbol m}_n
\right)
\rightsquigarrow
\boldsymbol G,
\end{equation}
where $\boldsymbol G$ is a centered Gaussian vector
\citep{vanderVaartWellner1996}.

The sign of the correction matters: the one-sided lower bound requires the upper quantile of $\max_\pi(\widehat m_n(\pi)-\widehat m_n^\ast(\pi))$. Since $-\boldsymbol G$ has the same distribution as $\boldsymbol G$ for a centered Gaussian vector, the
conditional distribution of
\begin{equation}
\sqrt n
\max_{\pi\in\mathcal P}
\left\{
\widehat m_n(\pi)
-
\widehat m_n^\ast(\pi)
\right\}
\end{equation}
consistently estimates the distribution of
\begin{equation}
\sqrt n
\max_{\pi\in\mathcal P}
\left\{
\widehat m_n(\pi)
-
m(\pi)
\right\}.
\end{equation}
Quantile consistency and continuity at the target quantile give the stated
simultaneous lower-band result.
\end{proof}

\begin{corollary}[Data-dependent commitment]
\label{cor:marcs_selection}
Let $\widehat\pi$ be any data-dependent selector satisfying
\[
\widehat\pi
\in
\mathcal B_{\rm S}(\delta)
\]
whenever that set is nonempty. Under the conditions of
Theorem~\ref{thm:marcs_fixed_menu},
\begin{equation}
\begin{aligned}
\liminf_{n\rightarrow\infty}
\Pr\Bigl\{
&
\mathcal B_{\rm S}(\delta)
=
\varnothing
\\
&
\text{or }
m(\widehat\pi)
\ge
\delta
\Bigr\}
\ge
1-\alpha.
\end{aligned}
\end{equation}
\end{corollary}

\begin{proof}
On the simultaneous event in
Theorem~\ref{thm:marcs_fixed_menu}, every member of
$\mathcal B_{\rm S}(\delta)$ has population joint margin at least
$\delta$. The conclusion therefore holds for any subsequent cost- or
margin-based selector from that set.
\end{proof}

\begin{remark}[Calibration-derived menus]
The theorem requires the menu and its policy parameters to be fixed
independently of the audit sample. In the reported experiments, the policy menu and its parameters are frozen using construction/selection data before the independent commitment audit. The theorem therefore aligns with the sample chronology conditional on this construction. Its remaining distributional and regularity conditions, and the finite Monte Carlo approximation, continue to limit any general claim of certification.
\end{remark}

\begin{remark}[Tied active constraints]
The separation condition excludes policies whose population risk and coverage
margins are exactly tied. At such kink points, the minimum map is only
directionally differentiable, and the ordinary bootstrap need not be
consistent without an additional correction
\citep{fang2019inference}.
\end{remark}

\begin{remark}[Zero accepted observations and bootstrap centering]
The positive population coverage requirement is asymptotic. A finite audit or bootstrap replicate may nevertheless accept no examples; the frozen implementation assigns an infeasible/undefined-risk outcome rather than fabricating a zero risk. Under a fixed positive coverage floor, the probability of such a replicate vanishes as the audit sample grows. The theorem concerns a centered common-resample maximum over the \emph{fixed} menu; separate per-policy marginal quantiles would not imply the stated selection corollary.
\end{remark}

\begin{remark}[Finite bootstrap budget]
The theorem concerns the ideal conditional bootstrap quantile, or a Monte
Carlo approximation with an increasing replicate budget. The canonical reported implementation uses $B=1000$ replicates. The frozen $B=100$ comparison assesses numerical stabilization, not finite-sample validity.
\end{remark}

\section{Complete RouterBench Operating Results}\label{app:routergrid}
\begin{figure*}[!tp]
\centering\includegraphics[width=.94\textwidth]{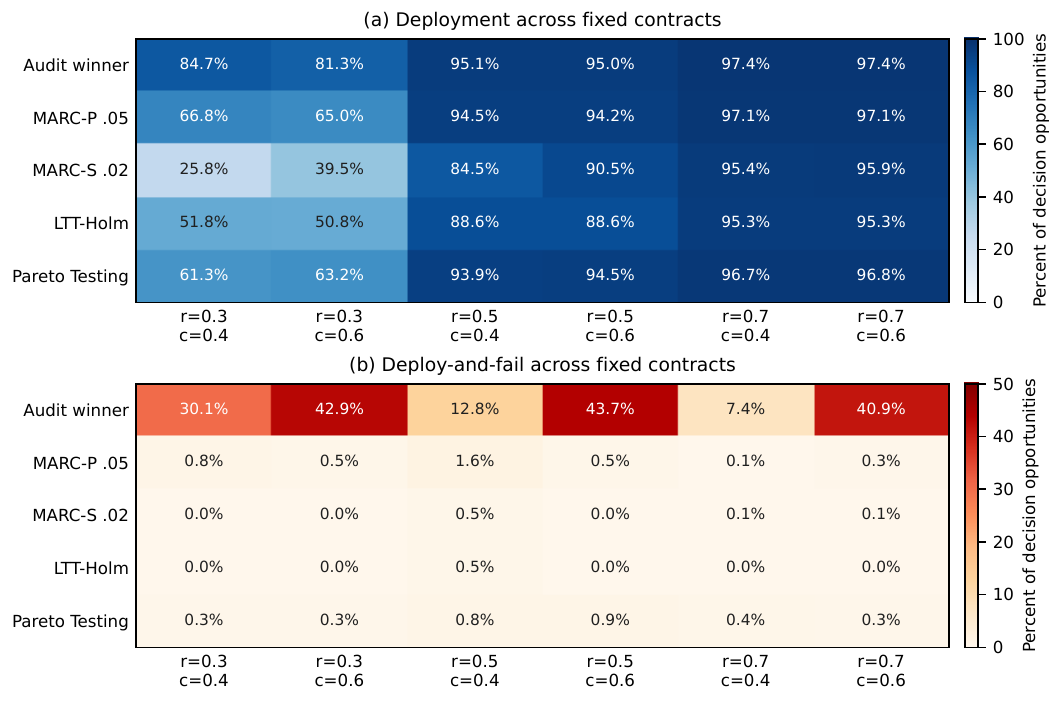}
\caption{RouterBench operating rates across the six prespecified risk--coverage contracts.}
\label{fig:appcontracts}
\end{figure*}

Full-grid dataset-equal deployment, unconditional deploy-and-fail, and conditional failure are given in Table~\ref{tab:appfull}. The table retains all frozen fixed-margin operating points, rather than selecting the most favorable test outcome. Figure~\ref{fig:appcontracts} shows six-contract heterogeneity for representative procedures. Tables~\ref{tab:full_mean_cost} and \ref{tab:full_tail_cost} separate the two cost objectives. Serving costs are not unconditional system costs.
\begin{table}[!htbp]
\centering
\footnotesize
\setlength{\tabcolsep}{3.3pt}
\begin{tabular}{lrrr}
\toprule
Procedure & Deploy & D\&F & Fail$|$D\\
\midrule
Audit winner & 91.82 & 29.63 & 32.22\\
Independent gate & 63.00 & 12.38 & 21.08\\
Fixed .00 & 91.82 & 22.25 & 24.40\\
Fixed .02 & 89.89 & 14.04 & 15.86\\
Fixed .05 & 87.72 & 6.06 & 7.25\\
Fixed .08 & 86.93 & 2.66 & 3.32\\
Fixed .10 & 86.32 & 1.43 & 1.78\\
Fixed .15 & 82.74 & 0.29 & 0.38\\
MARC-P .05 & 85.79 & 0.63 & 0.73\\
MARC-S .02 & 71.93 & 0.13 & 0.21\\
MARC-S .05 & 68.03 & 0.00 & 0 obs.\\
LTT-Holm & 78.38 & 0.09 & 0.12\\
Pareto & 84.41 & 0.49 & 0.60\\
\bottomrule
\end{tabular}
\caption{Full-grid RouterBench results (dataset-equal percent; $\alpha=.10$ when applicable).}
\label{tab:appfull}
\end{table}
\begin{table*}[!tp]\centering\footnotesize\setlength{\tabcolsep}{6pt}
\begin{tabular}{lrrrr}\toprule
Procedure & Deploy (\%) & D\&F (\%) & Fail$\mid$D (\%) & Selected cost (trace units/query) \\
\midrule
Audit winner & 91.82 & 29.52 & 32.14 & 0.000438 \\
Independent gate & 62.98 & 12.43 & 21.28 & 0.000328 \\
Fixed $.00$ & 91.82 & 29.17 & 31.78 & 0.000448 \\
Fixed $.02$ & 89.89 & 18.33 & 20.46 & 0.000431 \\
Fixed $.05$ & 87.72 & 7.94 & 9.36 & 0.000592 \\
Fixed $.08$ & 86.93 & 3.33 & 4.08 & 0.000808 \\
Fixed $.10$ & 86.32 & 1.73 & 2.12 & 0.001014 \\
Fixed $.15$ & 82.74 & 0.37 & 0.47 & 0.001455 \\
MARC-P $.00$ & 87.26 & 7.06 & 8.36 & 0.000549 \\
MARC-P $.02$ & 86.71 & 2.92 & 3.46 & 0.000734 \\
MARC-P $.05$ & 85.79 & 0.77 & 0.89 & 0.001071 \\
MARC-S $.00$ & 75.48 & 0.37 & 0.47 & 0.001124 \\
MARC-S $.02$ & 71.93 & 0.15 & 0.24 & 0.001125 \\
MARC-S $.05$ & 68.03 & 0.00 & 0.00 & 0.001130 \\
LTT-Bonferroni & 78.38 & 0.09 & 0.12 & 0.001392 \\
LTT-Holm & 78.38 & 0.09 & 0.12 & 0.001391 \\
Pareto Testing & 84.41 & 0.53 & 0.66 & 0.001297 \\
\bottomrule\end{tabular}\caption{RouterBench mean-cost optimization: full configuration grid (dataset-equal rates across six contracts and applicable score channels). Cost per incoming query, conditional on commitment.}\label{tab:full_mean_cost}\end{table*}
\begin{table*}[!tp]\centering\footnotesize\setlength{\tabcolsep}{6pt}
\begin{tabular}{lrrrr}\toprule
Procedure & Deploy (\%) & D\&F (\%) & Fail$\mid$D (\%) & Selected cost (trace units/query) \\
\midrule
Audit winner & 91.82 & 29.74 & 32.30 & 0.000749 \\
Independent gate & 63.03 & 12.32 & 20.91 & 0.000554 \\
Fixed $.00$ & 91.82 & 15.33 & 17.01 & 0.000833 \\
Fixed $.02$ & 89.89 & 9.74 & 11.25 & 0.000763 \\
Fixed $.05$ & 87.72 & 4.19 & 5.14 & 0.000978 \\
Fixed $.08$ & 86.93 & 2.00 & 2.55 & 0.001238 \\
Fixed $.10$ & 86.32 & 1.12 & 1.43 & 0.001449 \\
Fixed $.15$ & 82.74 & 0.20 & 0.29 & 0.001910 \\
MARC-P $.00$ & 87.26 & 2.96 & 3.81 & 0.000853 \\
MARC-P $.02$ & 86.71 & 1.36 & 1.74 & 0.001118 \\
MARC-P $.05$ & 85.79 & 0.48 & 0.57 & 0.001511 \\
MARC-S $.00$ & 75.48 & 0.20 & 0.28 & 0.001552 \\
MARC-S $.02$ & 71.93 & 0.11 & 0.18 & 0.001516 \\
MARC-S $.05$ & 68.03 & 0.00 & 0.00 & 0.001483 \\
LTT-Bonferroni & 78.38 & 0.09 & 0.12 & 0.001835 \\
LTT-Holm & 78.38 & 0.09 & 0.12 & 0.001835 \\
Pareto Testing & 84.41 & 0.46 & 0.54 & 0.001708 \\
\bottomrule\end{tabular}\caption{RouterBench tail-cost optimization: full configuration grid (dataset-equal rates across six contracts and applicable score channels). Cost per incoming query, conditional on commitment.}\label{tab:full_tail_cost}\end{table*}
\section{Exact-Control Applicability}\label{app:exact}
The exact-binomial rows use only the Bernoulli correctness channel. They are not comparable to all-channel entries in Table~\ref{tab:primary} as if the sample sizes were identical. Fractional/bounded correctness requires the appropriate bounded-loss tests rather than an exact binomial model. Table~\ref{tab:exact} compares all applicable methods on this common binary-outcome domain.
\begin{table*}[!tp]
\centering\small
\setlength{\tabcolsep}{5pt}
\begin{tabular}{lrrrrr}
\toprule
Procedure & Deploy (\%) & D\&F (\%) & Fail$\mid$D (\%) & Deployments & Failures \\
\midrule
Audit winner & 88.64 & 28.86 & 33.81 & 2,021 & 658 \\
Fixed $.05$ & 85.09 & 7.94 & 10.16 & 1,940 & 181 \\
MARC-P $.05$ & 83.07 & 0.79 & 1.04 & 1,894 & 18 \\
MARC-S $.02$ & 68.82 & 0.18 & 0.50 & 1,569 & 4 \\
MARC-S $.05$ & 64.69 & 0.00 & 0.00 & 1,475 & 0 \\
LTT-Bonf. & 75.26 & 0.09 & 0.13 & 1,716 & 2 \\
LTT-Holm & 75.26 & 0.09 & 0.13 & 1,716 & 2 \\
Pareto Testing & 81.58 & 0.57 & 0.95 & 1,860 & 13 \\
Exact-Bonf. & 77.72 & 0.09 & 0.12 & 1,772 & 2 \\
Exact-Holm & 77.72 & 0.09 & 0.12 & 1,772 & 2 \\
\bottomrule
\end{tabular}
\caption{Binary-score RouterBench comparison under the mean-cost objective (dataset-equal rates; pooled counts; nominal $\alpha=.10$).}
\label{tab:exact}
\end{table*}

\section{Failure Magnitude and Dataset-Level Inference}\label{app:failure}
\begin{figure}[!htbp]
\centering\includegraphics[width=.99\columnwidth]{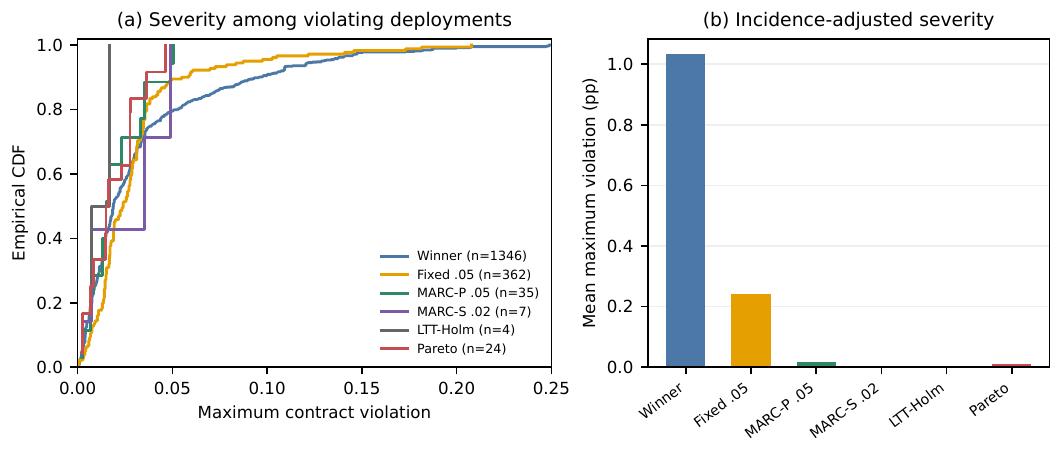}
\caption{Distribution of held-out contract-violation severity among failed commitments.}
\label{fig:appseverity}
\end{figure}

\paragraph{Estimands and aggregation.} For each dataset, unconditional deploy-and-fail is the fraction of decision opportunities committing a policy that violates at least one fresh contract constraint. Conditional failure is evaluated among committed opportunities, then the available dataset-level conditional proportions are equally weighted. Accordingly, the product of the separately equal-weighted deployment and conditional-failure columns need not equal equal-weighted deploy-and-fail. When a method makes no commitments on a dataset, its conditional failure is missing, not zero. Each cluster bootstrap resamples datasets, rather than treating within-dataset seed/contract cells as independent. Intervals describe dataset heterogeneity, not a finite-sample bound on an unobserved policy's true risk.
\paragraph{Severity and aggregation.} Eq.~\ref{eq:severity} measures the maximum of positive risk excess and coverage shortfall, assigning zero to nonviolating commitments. Incidence-adjusted severity averages over the stated evaluation population; conditional severity averages over violations only. The latter may increase even when a commitment rule reduces failure exposure. Figure~\ref{fig:appseverity} displays severity distributions, and Table~\ref{tab:violtypes} separates risk-only, coverage-only, and joint failures. The RouterBench audit winner has conditional mean, median, and p90 severity of 3.15, 1.94, and 7.39 percentage points, respectively. Table~\ref{tab:appstrict} reports a prespecified strict contract, and Table~\ref{tab:appdataset} presents dataset-level operating rates. Dataset-cluster uncertainty (Table~\ref{tab:inference}) and leave-one-dataset-out sensitivity (Table~\ref{tab:v3_lodo}) respect the 19 task-family reporting units.
\begin{table}[!htbp]\centering\small\setlength{\tabcolsep}{3.5pt}
\begin{tabular}{lrrrr}\toprule
Procedure & Risk only & Cov. only & Both & D\&F \\
\midrule
Audit winner & 10.83 & 16.64 & 2.15 & 29.63 \\
Independent gate & 2.85 & 9.22 & 0.31 & 12.38 \\
Fixed $.05$ & 3.42 & 2.52 & 0.12 & 6.06 \\
MARC-P $.05$ & 0.54 & 0.09 & 0.00 & 0.63 \\
MARC-S $.02$ & 0.11 & 0.02 & 0.00 & 0.13 \\
LTT-Holm & 0.09 & 0.00 & 0.00 & 0.09 \\
Pareto Testing & 0.27 & 0.22 & 0.00 & 0.49 \\
\bottomrule\end{tabular}
\caption{Mutually exclusive risk-only, coverage-only, and joint held-out contract violations, as a percentage of all RouterBench decision opportunities, dataset-equal. The components sum to deploy-and-fail before display rounding.}
\label{tab:violtypes}\end{table}

\begin{table}[!htbp]
\centering
\footnotesize
\setlength{\tabcolsep}{3.3pt}
\begin{tabular}{lrrr}
\toprule
Procedure & Deploy & D\&F & Fail/dep.\\
\midrule
Audit winner & 81.32 & 42.89 & 652/1236\\
LTT-Holm & 50.79 & 0.00 & 0/772\\
MARC-P .05 & 65.00 & 0.53 & 8/988\\
MARC-S .02 & 39.47 & 0.00 & 0/600\\
Pareto & 63.16 & 0.26 & 4/960\\
\bottomrule
\end{tabular}
\caption{RouterBench strict $(r^\star,c^\star)=(.30,.60)$ contract. Dataset-equal rates and observed counts use distinct denominators.}
\label{tab:appstrict}
\end{table}
\begin{table}[!htbp]
\centering
\scriptsize
\setlength{\tabcolsep}{2.9pt}
\begin{tabular}{lrrrr}
\toprule
Dataset & Win D & Win D\&F & P D & P D\&F\\
\midrule
arc-challenge & 100.00 & 30.00 & 100.00 & 0.00\\
chinese\_zodiac & 85.83 & 42.50 & 65.83 & 0.00\\
consensus\_summary & 64.58 & 29.17 & 48.33 & 0.21\\
grade-school-math & 33.33 & 4.17 & 33.33 & 0.00\\
hellaswag & 100.00 & 20.00 & 100.00 & 0.00\\
mbpp & 90.83 & 47.50 & 66.67 & 1.67\\
mmlu-elementary-mathematics & 90.00 & 66.25 & 65.00 & 0.83\\
mmlu-high-school-biology & 100.00 & 30.42 & 100.00 & 0.83\\
mmlu-high-school-macroeconomics & 100.00 & 32.08 & 100.00 & 2.50\\
mmlu-high-school-psychology & 100.00 & 25.42 & 100.00 & 3.33\\
mmlu-miscellaneous & 100.00 & 17.50 & 100.00 & 0.00\\
mmlu-moral-disputes & 100.00 & 19.58 & 91.67 & 0.83\\
mmlu-moral-scenarios & 100.00 & 32.08 & 100.00 & 0.00\\
mmlu-nutrition & 100.00 & 30.83 & 100.00 & 0.83\\
mmlu-philosophy & 100.00 & 34.17 & 93.33 & 0.00\\
mmlu-prehistory & 100.00 & 22.08 & 100.00 & 0.00\\
mmlu-professional-law & 80.00 & 22.92 & 66.67 & 0.83\\
mmlu-professional-psychology & 100.00 & 26.67 & 99.17 & 0.00\\
winogrande & 100.00 & 29.58 & 100.00 & 0.00\\
\bottomrule
\end{tabular}
\caption{Full-grid per-dataset deployment and deploy-and-fail percentages (winner versus MARC-P), each dataset weighted equally in primary aggregates.}
\label{tab:appdataset}
\end{table}
\begin{table*}[!tp]\centering\footnotesize\setlength{\tabcolsep}{5pt}
\begin{tabular}{lrrr}\toprule
Procedure & Deploy (\%) & D\&F (\%) & Severity (pp) \\
\midrule
Audit winner & 91.82 [83.46, 98.03] & 29.63 [24.34, 35.53] & 1.03 [0.70, 1.38] \\
MARC-P $.05$ & 85.79 [75.92, 94.08] & 0.63 [0.26, 1.06] & 0.01 [0.01, 0.02] \\
MARC-S $.02$ & 71.93 [62.89, 80.79] & 0.13 [0.02, 0.26] & 0.00 [0.00, 0.01] \\
MARC-S $.05$ & 68.03 [58.55, 77.28] & 0.00 [0.00, 0.00] & 0.00 [0.00, 0.00] \\
LTT-Holm & 78.38 [67.85, 88.20] & 0.09 [0.00, 0.22] & 0.00 [0.00, 0.00] \\
Pareto Testing & 84.41 [74.85, 92.76] & 0.49 [0.16, 0.92] & 0.01 [0.00, 0.02] \\
Fixed $.10$ & 86.32 [76.40, 94.78] & 1.43 [0.81, 2.11] & 0.04 [0.02, 0.06] \\
\bottomrule\end{tabular}
\caption{RouterBench dataset-bootstrap 95\% intervals over 19 task families. Severity is the maximum contract violation, with zero assigned to nonviolations.}\label{tab:inference}\end{table*}

\begin{table}[!htbp]\centering\footnotesize\setlength{\tabcolsep}{4pt}
\begin{tabular}{llrr}\toprule
Method vs winner & Outcome & LODO min & LODO max \\
\midrule
MARC-P .05 & D\&F & -30.38 & -26.98 \\
MARC-P .05 & Deploy & -6.37 & -4.98 \\
Fixed .10 & D\&F & -29.54 & -26.39 \\
Fixed .10 & Deploy & -5.81 & -4.51 \\
Pareto & D\&F & -30.52 & -27.07 \\
Pareto & Deploy & -7.82 & -6.34 \\
LTT-Holm & D\&F & -30.95 & -27.50 \\
LTT-Holm & Deploy & -14.19 & -12.11 \\
MARC-S .02 & D\&F & -30.90 & -27.45 \\
MARC-S .02 & Deploy & -21.00 & -18.59 \\
\bottomrule\end{tabular}
\caption{Leave-one-dataset-out ranges for paired method-minus-winner differences (percentage points; 19 omissions).}\label{tab:v3_lodo}\end{table}

\paragraph{Dataset-level paired effects.} Table~\ref{tab:v3_lodo} summarizes the already-frozen leave-one-dataset-out estimator for method-minus-winner effects. For example, MARC-P's deploy-and-fail reduction relative to the winner remains between $-30.38$ and $-26.98$ percentage points across the 19 possible omissions; its corresponding deployment difference ranges from $-6.37$ to $-4.98$ points. These are ranges of descriptive benchmark-panel contrasts, not finite-sample confidence bounds or independent replicate tests. Intervals for failure-conditional quantities should be interpreted only when deployment denominators are positive.
\section{Nested Audit-Size Robustness}\label{app:audit}
\paragraph{Paired comparison and uncertainty.} The independent audit grows through nested frozen query-ID hash prefixes, keeping the construction and test samples constant for all audit sizes within each seed. We contrast A40 with A10 on each task and split and aggregate the changes at the dataset level; marginal pointwise error bars alone do not characterize this paired change. Frozen outputs include per-dataset paired results, retained/deferred decisions, supported-set counts, and costs on common deployment opportunities. The reported audit-size comparison keeps buffer settings constant across audit sizes.

\begin{figure*}[!tp]
\centering\includegraphics[width=.94\textwidth]{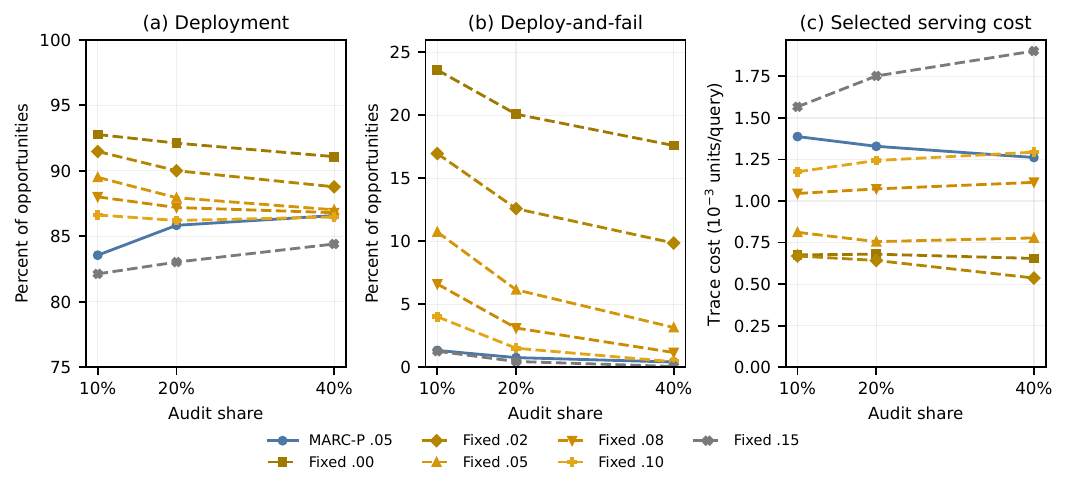}
\caption{Frozen audit-size results for MARC-P and fixed margins. Fixed $.15$ also gains deployment as audit evidence grows.}
\label{fig:appaudit}
\end{figure*}

The pre-frozen robustness design fixes a 20\% construction sample, 40\% audit reservoir, and 40\% common test. A deterministic query-ID hash gives $A_{10}\subset A_{20}\subset A_{40}$. At each audit size, the same frozen candidate menus and numerical buffers are used. Figure~\ref{fig:appaudit} displays the point-margin sweep and MARC-P; Table~\ref{tab:appallmargin} lists all audit-size endpoints, and Tables~\ref{tab:auditpaired} and \ref{tab:paired_intervals} report paired changes and dataset intervals. The test population is common \emph{within the audit-size experiment}, but differs from the primary 20/20/60 RouterBench evaluation.
\begin{table*}[!tp]
\centering
\footnotesize
\setlength{\tabcolsep}{3.3pt}
\begin{tabular}{lrrrrrr}
\toprule
Procedure & D10 & D20 & D40 & DF10 & DF20 & DF40\\
\midrule
MARC-P .05 & 83.55 & 85.83 & 86.58 & 1.34 & 0.77 & 0.42\\
Fixed 0.00 & 92.76 & 92.11 & 91.07 & 23.60 & 20.09 & 17.60\\
Fixed 0.02 & 91.47 & 90.00 & 88.77 & 16.96 & 12.59 & 9.86\\
Fixed 0.05 & 89.50 & 87.94 & 87.02 & 10.73 & 6.15 & 3.16\\
Fixed 0.08 & 88.00 & 87.19 & 86.80 & 6.61 & 3.11 & 1.17\\
Fixed 0.10 & 86.62 & 86.21 & 86.47 & 4.01 & 1.51 & 0.44\\
Fixed 0.15 & 82.13 & 83.03 & 84.41 & 1.27 & 0.46 & 0.07\\
\bottomrule
\end{tabular}
\caption{Nested-audit deployment ($D$) and deploy-and-fail ($DF$) rates at 10/20/40\% audit; fixed method settings.}
\label{tab:appallmargin}
\end{table*}
Matched-common-deployment cost comparisons from the frozen paired ledger avoid conflating policy selection with changing deployment populations. The existing evidence establishes the direction of the MARC-P cost change but not an overall service optimum without accounting for fallback and deferral. The dataset-specific A10-to-A40 changes are heterogeneous; the average should not be read as a monotonic claim for every task.
\begin{table*}[!tp]\centering\small\setlength{\tabcolsep}{4pt}
\begin{tabular}{lrrrrr}\toprule
Procedure & $\Delta$Deploy (pp) & $\Delta$D\&F (pp) & $\Delta$Severity (pp) & $\Delta$ common cost & $\Delta$ supported-set size \\
\midrule
MARC-P $.05$ & +3.03 & -0.92 & -0.030 & -0.000193 & +1.49 \\
Fixed $.05$ & -2.48 & -7.58 & -0.356 & +0.000082 & -0.57 \\
Fixed $.10$ & -0.15 & -3.57 & -0.147 & +0.000129 & -0.65 \\
Fixed $.15$ & +2.28 & -1.21 & -0.044 & +0.000198 & +0.15 \\
\bottomrule\end{tabular}\caption{Nested-audit paired A40--A10 changes (dataset-equal). Serving-cost differences use common deployments.}\label{tab:auditpaired}\end{table*}
\begin{table}[!htbp]\centering\scriptsize\setlength{\tabcolsep}{2pt}
\begin{tabular}{lrrr}\toprule
A40--A10 & Mean & 95\% CI & $+/-/0$ tasks \\
\midrule
Deployment (pp) & +3.03 & [+1.01, +5.44] & 9/2/8 \\
Deploy-and-fail (pp) & -0.92 & [-1.67, -0.22] & 3/10/6 \\
Supported-set size & +1.49 & [+1.04, +1.95] & 17/2/0 \\
\bottomrule\end{tabular}\caption{MARC-P $.05$ paired A40--A10 changes over 19 RouterBench datasets. Descriptive dataset-cluster percentile intervals use 20,000 bootstrap resamples of the paired results.}\label{tab:paired_intervals}\end{table}
\section{Learned Semantic-Router Evaluation}\label{app:learned}

\paragraph{Learned policy construction.} Query embeddings enter the CARROT-style nearest-neighbor quality estimator. Frozen model-cost preferences and acceptance thresholds give 48 routing policies per dataset. These policies are fitted and selected before the independent commitment audit; no learned parameters are updated using audit or test outcomes.

The learned panel covers 10 datasets, 11 models, and 20 seeds with independent 40/20/20/20 fit/selection/audit/test splits. Query traces, embeddings, and split partitions were verified against reference records; deterministic replay and repeated commitment runs agreed. The evaluation isolates commitment behavior for an independently trained routing family.

\begin{figure*}[!tp]
\centering\includegraphics[width=.94\textwidth]{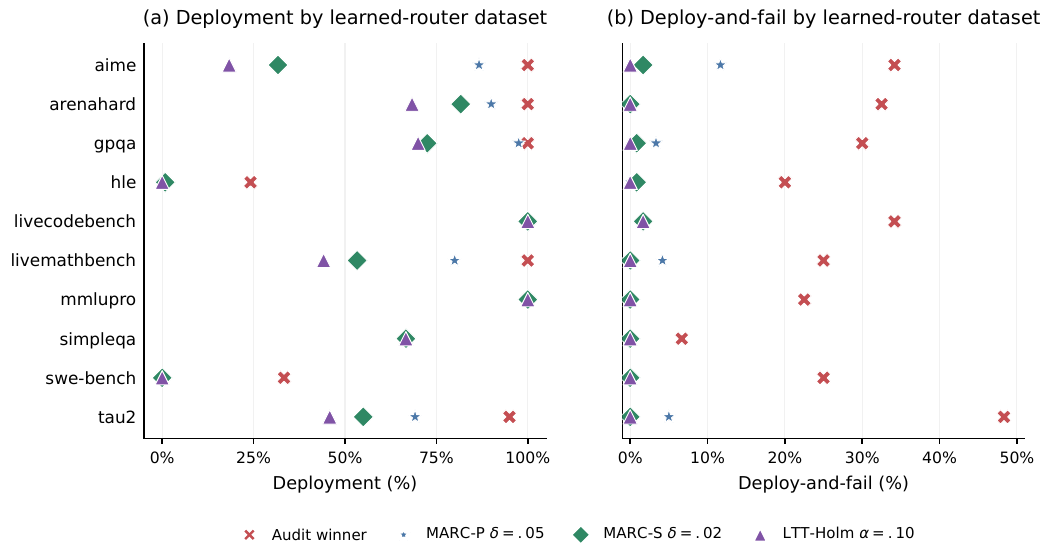}
\caption{Learned-router deployment and deploy-and-fail by dataset, using the frozen KNN menu.}
\label{fig:applearnedhet}
\end{figure*}

Main-text Table~\ref{tab:learned} reports the full-grid and strict-contract comparisons on pooled decision opportunities with observed failure/deployment counts. Figure~\ref{fig:applearnedhet} shows the heterogeneity that pooled failure rates can hide; Tables~\ref{tab:learneddata} and \ref{tab:learned_full_dataset} give per-dataset operating rates, and Table~\ref{tab:v3_learned_counts} gives actual deployment/failure counts. Conditional failure is undefined when a dataset has zero deployments, rather than imputed to zero. At the strict contract, stronger formal and simultaneous methods observe no failures but often deploy in only about one fifth to one quarter of opportunities. Residual MARC-P failures need not be less severe conditional on failure even when total violation incidence drops.
\begin{table}[!htbp]
\centering
\scriptsize
\setlength{\tabcolsep}{3.3pt}
\begin{tabular}{lrrrr}
\toprule
Dataset & Win D & Win DF & P D & P DF\\
\midrule
aime & 100.00 & 34.17 & 86.67 & 11.67\\
arenahard & 100.00 & 32.50 & 90.00 & 0.00\\
gpqa & 100.00 & 30.00 & 97.50 & 3.33\\
hle & 24.17 & 20.00 & 0.00 & 0.00\\
livecodebench & 100.00 & 34.17 & 100.00 & 1.67\\
livemathbench & 100.00 & 25.00 & 80.00 & 4.17\\
mmlupro & 100.00 & 22.50 & 100.00 & 0.00\\
simpleqa & 66.67 & 6.67 & 66.67 & 0.00\\
swe-bench & 33.33 & 25.00 & 0.00 & 0.00\\
tau2 & 95.00 & 48.33 & 69.17 & 5.00\\
\bottomrule
\end{tabular}
\caption{Learned-router deployment and unconditional deploy-and-fail by dataset (percent; pooled within task).}
\label{tab:learneddata}
\end{table}
\begin{table*}[!tp]\centering\scriptsize\setlength{\tabcolsep}{4pt}
\begin{tabular}{lrrrrrr}
\toprule
Dataset & Winner D & Winner D\&F & MARC-P D & MARC-P D\&F & LTT D & Pareto D \\
\midrule
aime & 100.0 & 34.2 & 86.7 & 11.7 & 18.3 & 48.3 \\
arenahard & 100.0 & 32.5 & 90.0 & 0.0 & 68.3 & 77.5 \\
gpqa & 100.0 & 30.0 & 97.5 & 3.3 & 70.0 & 73.3 \\
hle & 24.2 & 20.0 & 0.0 & 0.0 & 0.0 & 0.8 \\
livecodebench & 100.0 & 34.2 & 100.0 & 1.7 & 100.0 & 100.0 \\
livemathbench & 100.0 & 25.0 & 80.0 & 4.2 & 44.2 & 60.0 \\
mmlupro & 100.0 & 22.5 & 100.0 & 0.0 & 100.0 & 100.0 \\
simpleqa & 66.7 & 6.7 & 66.7 & 0.0 & 66.7 & 66.7 \\
swe-bench & 33.3 & 25.0 & 0.0 & 0.0 & 0.0 & 0.0 \\
tau2 & 95.0 & 48.3 & 69.2 & 5.0 & 45.8 & 55.8 \\
\bottomrule\end{tabular}\caption{Learned-router deployment ($D$) and deploy-and-fail (D\&F) by dataset, averaged over 20 seeds and six fixed contracts (percent).}\label{tab:learned_full_dataset}\end{table*}
\begin{table*}[!tp]\centering\footnotesize\setlength{\tabcolsep}{5pt}
\begin{tabular}{lrrrrr}\toprule
Task ($120$ opportunities each) & Winner & MARC-P $.05$ & MARC-S $.02$ & LTT-Holm & Pareto \\
\midrule
aime & 120/41 & 104/14 & 38/2 & 22/0 & 58/3 \\
arenahard & 120/39 & 108/0 & 98/0 & 82/0 & 93/0 \\
gpqa & 120/36 & 117/4 & 87/1 & 84/0 & 88/1 \\
hle & 29/24 & 0/0 & 1/1 & 0/0 & 1/1 \\
livecodebench & 120/41 & 120/2 & 120/2 & 120/2 & 120/1 \\
livemathbench & 120/30 & 96/5 & 64/0 & 53/0 & 72/0 \\
mmlupro & 120/27 & 120/0 & 120/0 & 120/0 & 120/0 \\
simpleqa & 80/8 & 80/0 & 80/0 & 80/0 & 80/0 \\
swe-bench & 40/30 & 0/0 & 0/0 & 0/0 & 0/0 \\
tau2 & 114/58 & 83/6 & 66/0 & 55/0 & 67/0 \\
\bottomrule\end{tabular}
\caption{Learned-router deployment/failure counts by dataset, over 20 seeds and six fixed contracts (mean-cost objective). A $0/0$ entry indicates no deployment.}\label{tab:v3_learned_counts}\end{table*}

\section{Serving-Cost Objectives and Sensitivity}\label{app:economics}
\paragraph{Three economic populations.} (i) Policy serving-cost $J(\pi)$ is computed over incoming queries, including abstention at the frozen zero policy-level cost; (ii) average $J$ \emph{among committed decision opportunities} summarizes the policy that a commitment rule selects, and is sensitive to which opportunities each method chooses to serve; (iii) a whole-system objective additionally requires an application-dependent cost for never deploying a policy, fallback, or delay. The frozen trace identifies the first two but not the third. Matched-deployment comparisons condition on opportunities common to a pair of rules or to two audit-size treatments and are presented as comparators, not complete system valuations. The mean-cost and tail-cost optimization objectives remain separate throughout.

The experiment has separate \texttt{mean\_cost} and \texttt{tail\_cost} optimization objectives. Summarizing selected policy cost over a mixture of these objectives is not a homogeneous monetary estimand. Figure~\ref{fig:frontier}(b) therefore uses an explicitly mean-cost-only RouterBench comparison; table~\ref{tab:objcost} separates objectives. Costs are conditional on deployment, so methods choosing different cells cannot be ranked by complete end-to-end operational expense without a specified deferral/fallback price. Tables~\ref{tab:v3_matched_cost} and \ref{tab:v3_cost_inference} report matched and objective-specific cost uncertainty; Table~\ref{tab:v3_utility} gives the frozen illustrative penalty scenarios. These do not identify an overall optimum without real deferral and fallback costs.
\begin{table}[!htbp]\centering\small\setlength{\tabcolsep}{4pt}
\begin{tabular}{lrr}\toprule
Procedure & Mean objective & Tail objective \\
\midrule
Audit winner & 0.000438 & 0.000749 \\
Fixed $.05$ & 0.000592 & 0.000978 \\
Fixed $.10$ & 0.001014 & 0.001449 \\
Fixed $.15$ & 0.001455 & 0.001910 \\
MARC-P $.05$ & 0.001071 & 0.001511 \\
MARC-S $.02$ & 0.001125 & 0.001516 \\
MARC-S $.05$ & 0.001130 & 0.001483 \\
Pareto Testing & 0.001297 & 0.001708 \\
LTT-Holm & 0.001391 & 0.001835 \\
\bottomrule\end{tabular}\caption{RouterBench selected-policy cost by optimization objective (trace units per incoming query; dataset-equal, conditional on commitment).}\label{tab:objcost}\end{table}

\begin{table*}[!tp]\centering\scriptsize\setlength{\tabcolsep}{3pt}
\begin{tabular}{llrrl}\toprule
Objective & Comparison (MARC-P $.05$ vs.) & Shared $N$ & $\Delta J$ & 95\% cluster interval \\
\midrule
Mean & Winner & 3912 & +0.000844 & [+0.000408, +0.001444] \\
Mean & Fixed .10 & 3912 & +0.000070 & [-0.000019, +0.000159] \\
Mean & Pareto & 3841 & -0.000243 & [-0.000429, -0.000069] \\
Mean & LTT-Holm & 3574 & -0.000517 & [-0.000770, -0.000295] \\
Tail & Winner & 3912 & +0.001105 & [+0.000495, +0.002032] \\
Tail & Fixed .10 & 3912 & +0.000082 & [-0.000016, +0.000187] \\
Tail & Pareto & 3841 & -0.000224 & [-0.000461, -0.000009] \\
Tail & LTT-Holm & 3574 & -0.000571 & [-0.000903, -0.000314] \\
\bottomrule\end{tabular}
\caption{Matched RouterBench serving-cost differences for MARC-P versus each comparator on their jointly deployed opportunities. Dataset-equal means and 95\% dataset-bootstrap intervals; positive $\Delta J$ indicates higher MARC-P cost. Comparator rows use distinct matched populations.}\label{tab:v3_matched_cost}\end{table*}

\begin{table}[!htbp]\centering\footnotesize\setlength{\tabcolsep}{3pt}
\begin{tabular}{llrl}\toprule
Objective & Method & Cost & 95\% dataset interval \\
\midrule
Mean & Winner & 0.000438 & [0.000185, 0.000752] \\
Mean & MARC-P .05 & 0.001071 & [0.000563, 0.001750] \\
Mean & Pareto & 0.001297 & [0.000781, 0.001968] \\
Mean & LTT-Holm & 0.001391 & [0.000774, 0.002177] \\
Tail & Winner & 0.000749 & [0.000319, 0.001283] \\
Tail & MARC-P .05 & 0.001511 & [0.000749, 0.002620] \\
Tail & Pareto & 0.001708 & [0.000971, 0.002695] \\
Tail & LTT-Holm & 0.001835 & [0.000993, 0.003000] \\
\bottomrule\end{tabular}
\caption{Selected-policy serving cost by optimization objective, conditional on deployment. Dataset-equal means and descriptive 95\% bootstrap intervals over 19 tasks.}\label{tab:v3_cost_inference}\end{table}

\begin{table*}[!tp]\centering\footnotesize\setlength{\tabcolsep}{6pt}
\begin{tabular}{rrrrrr}\toprule
Failure penalty & Defer penalty & Winner & MARC-P $.05$ & Pareto & LTT-Holm \\
\midrule
0.001 & 0.0001 & 0.000696 & 0.000892 & 0.001083 & 0.001094 \\
0.001 & 0.01 & 0.001506 & 0.002299 & 0.002626 & 0.003235 \\
0.1 & 0.0001 & 0.029918 & 0.001652 & 0.001604 & 0.001181 \\
0.1 & 0.01 & 0.030728 & 0.003059 & 0.003147 & 0.003322 \\
\bottomrule\end{tabular}
\caption{Illustrative mean-cost decision penalties over the fixed $4\times4$ failure/deferral grid. The scenario objective is $\mathbb E[D J_{\rm trace}+\lambda_F\mathbf{1}\{D\cap F\}+\lambda_D(1-D)]$; penalty values are hypothetical.}\label{tab:v3_utility}\end{table*}

\paragraph{What matched comparisons identify.} In Table~\ref{tab:v3_matched_cost}, a method pair contributes only decisions on which \emph{both} rules deploy. This isolates the serving-cost difference between their selected policies on the same fixed opportunities; it does not estimate the cost of deferring other opportunities. Objective-separated 95\% dataset-bootstrap intervals are descriptive and not extrapolations to every future traffic distribution. The frozen scenario table reports a transparent linear penalty functional on the \emph{mean-cost} objective using only the existing $(\lambda_F,\lambda_D)$ grid, and is not a business-calibrated monetary return or a new parameter search. Scenarios cannot justify a universal optimum without a real operator's failure, fallback and lost-service values.
\section{Operating-Point and Monte Carlo Sensitivity}\label{app:sensitivity}
Table~\ref{tab:appfull} includes the full prespecified MARC buffer and fixed-margin sweeps. RouterBench MARC-P and MARC-S were recomputed with $B=1000$ using the verified fixed implementation. Relative to $B=100$, MARC-P $.05$ deployment moved 85.44\% to 85.79\%; MARC-S $.02$ moved 71.69\% to 71.93\%, and MARC-S $.05$ moved 67.83\% to 68.03\%. The latter still had zero observed failures. These changes reflect Monte Carlo stabilization at fixed operating settings. Table~\ref{tab:v3_alpha_sensitivity} additionally reports the existing formal-control $\alpha=.05$ sensitivity; MARC settings remain unchanged.

\begin{table}[!htbp]\centering\footnotesize\setlength{\tabcolsep}{5pt}
\begin{tabular}{lrrr}\toprule
Procedure & $\alpha$ & Deploy & D\&F \\
\midrule
LTT-Holm & 0.10 & 78.38 & 0.09 \\
LTT-Bonf. & 0.10 & 78.38 & 0.09 \\
Pareto & 0.10 & 84.41 & 0.49 \\
LTT-Holm & 0.05 & 76.75 & 0.00 \\
LTT-Bonf. & 0.05 & 76.75 & 0.00 \\
Pareto & 0.05 & 82.83 & 0.37 \\
\bottomrule\end{tabular}
\caption{RouterBench formal-control sensitivity (dataset-equal percent) at nominal $\alpha\in\{.10,.05\}$.}\label{tab:v3_alpha_sensitivity}\end{table}

\section{External Menu-Size Sensitivity}\label{app:boundary}
The retain/replace/defer comparison is reported in main-text Table~\ref{tab:actions}. Table~\ref{tab:xroute} provides a complementary xRouteBench menu-size trajectory: deployment is constant while observed held-out failure decreases as the menu grows. Together with the main RouterBench findings, this highlights the role of policy composition, feasibility margins, and cost-driven selection rather than menu cardinality alone.

\begin{table}[!htbp]
\centering
\footnotesize
\setlength{\tabcolsep}{3.3pt}
\begin{tabular}{lrrr}
\toprule
Menu size & Deploy & Fail$|$D & D\&F\\
\midrule
7 & 66.67 & 9.62 & 6.42\\
14 & 66.67 & 7.25 & 4.83\\
21 & 66.67 & 5.25 & 3.50\\
28 & 66.67 & 2.50 & 1.67\\
35 & 66.67 & 0.00 & 0.00\\
\bottomrule
\end{tabular}
\caption{xRouteBench held-out menu-size sensitivity (percent). Its policy menu and contract differ from the main RouterBench study.}
\label{tab:xroute}
\end{table}
\paragraph{Evaluation scope.} Independent-audit results exclude earlier same-split diagnostics. xRouteBench uses a separate frozen policy menu and contract. In every study, numerical service targets are fixed before auditing and held through selection and test evaluation.
\end{document}